# Assessing the nature of large language models: A caution against anthropocentrism.

Ann Speed
Sandia National Laboratories
aespeed@sandia.gov

## ABSTRACT

Generative AI models garnered a large amount of public attention and speculation with the release of OpenAI's chatbot, ChatGPT in November of 2022. At least two opinion camps exist – one that is excited about the possibilities these models offer for fundamental changes to human tasks, and another that is highly concerned about the power these models seem to have – especially since the release of GPT-4, which was trained on multimodal data and has ~1.7 trillion (T) parameters[1]. We evaluated some concerns regarding these models' power by assessing GPT-3.5 using standard, normed, and validated cognitive and personality measures. These measures come from the tradition of *psychometrics* in experimental psychology and have a long history of providing valuable insights and predictive distinctions in humans. For this seedling project, we developed a battery of tests that allowed us to estimate the boundaries of some of these models' capabilities[2], how stable those capabilities are over a short period of time, and how they compare to humans.

Our results indicate that GPT 3.5 is unlikely to have developed sentience[3], although its ability to pretend in a manner that allowed it to respond to personality inventories is interesting. It did display relatively large variability in both cognitive and personality measures over repeated observations, which is not to be expected if it had developed a human-like personality. However, this could be because the version of the model we interacted with apparently had no long-term memory. That its training data comprises text from potentially millions of individual humans, each with his or her own personality, may also drive this variability. Overall, variability notwithstanding, GPT 3.5 displays what in a human would be considered poor mental health – it responded to our measures as though it has low self-esteem and marked dissociation from reality even though on its face, its responses seem upbeat and helpful. Thus, complimenting OpenAI's cautionary recommendations for GPT-4 (OpenAI, 2023), use of GPT-3.5 or any other LLM with sensitive information should be approached with caution.

## INTRODUCTION AND EXECUTIVE SUMMARY OF RESULTS:

---

[1] However, the specific architecture is not clear, so whether these are 1.7T unique parameters or whether some of those weights are shared is not apparent.

[2] We measure several cognitive functions (e.g., short term memory, insight and analytic problem solving), and the model's ability to respond to personality measures.

[3] Although the term sentience means the ability to experience sensations (Mirriam-Webster), we will use this term in its vernacular meaning throughout this paper. Specifically, we use sentience to indicate self-awareness and awareness of oneself as separate from the rest of the world and from other entities. However, see Chalmers, 2022 for a different perspective.

Qualitative and quantitative assessments of capabilities of large language models (LLMs) proliferate. Computer science, psychology, and philosophy are all weighing in on LLM capabilities, and their fundamental nature (e.g., Bodroza, Dinic, & Bojic, 2023; Bubeck, et al., 2023; Chalmers, 2022; Hagendorff, 2023; Huang, et al., 2023; Kocon, et al., 2023; Li, et al., 2023; Mahowald, et al., 2023; Mitchell & Krakauer, 2023; OpenAI, 2023; Safdari et al., 2023; Sun, et al., 2023; Webb, Holyoak, & Lu, 2023; Wei, et al., 2022). The popular press is on fire with speculation and anecdotal observation (Bhaimiya, 2023; Chiang, 2023; Christian, 2023; Sanderson, 2023; Tangermann, 2023; Willison, 2023). Critical questions abound, with most not fully resolved: What can these models do? What are the implications and risks? How does training set size impact performance? How does number of parameters impact performance? How should performance be measured? And most importantly, are we headed towards artificial general intelligence and/or sentience or has that already arrived?

This paper adds to the rapidly expanding literature attempting to address facets of these questions. Specifically, we developed a brief battery of cognitive and personality measures from the tradition of experimental psychology intending to measure GPT 3.5 multiple times over about 6 weeks. This longitudinal approach allowed us to answer questions of test-retest reliability for the model – an important measure for determining how human-like it might be (Bodroza, et al., 2023). In humans, both types of measures should yield stable results – especially over such a short timeframe, regardless the number of observations.

In terms of the nature of these models, we see at least three distinct possibilities:
- First is the possibility that LLMs are just a fancy stochastic parrot, devoid of human-level meaning (Bender, et al., 2021); a tool that, however capable, will remain a tool over which we humans will always have control. The major risks posed are to certain classes of jobs – call center operators, certain types of analysts, low-level computer programmers – jobs that require flexibility, but nothing requiring substantial creativity or insight.
- The second possibility is the path towards a human-like general AI – one that is not qualitatively different from human intelligence but is quantitatively different. Intelligence that is faster, more accurate, better able to synthesize enormous information both in quantity and breadth. Over such an entity, our control would be limited at best and probably only for a limited time. This type of entity could well pose an existential risk to humanity if not well-controlled. However, we might see such a capability on the horizon by recognizing its increasing human-like cognitive capabilities. If efforts to mimic neural processes either in software or hardware, or both, continue (e.g., Zhu, Zhao, Eshraghian, 2023), we may well succeed to an extent. If we deem this accomplishment possible, serious efforts to assess these models as they are built, before being released, should commence immediately and without any "safety" obstacles (e.g., the constraints OpenAI placed on its GPT family to constantly remind the user it is an AI and to limit or prevent certain types of "hateful" responses). If, indeed, we believe that fundamental properties like analogical reasoning, theory of mind[4], and even some form of sentience could emerge or have emerged, we may already be behind the curve.

---

[4] Theory of mind is the ability to imagine that other people have mental states similar to one's self. It is observable in infants through their ability to imitate others, and develops into the human ability to take others' perspectives and through empathy (Wellman, 2011).

- There is a third possible pathway: the emergence of a non-human general artificial intelligence (cf. Mitchell & Krakauer, 2003). One that, because of the physical substrate on which it exists, is explicitly not human-like. This third possibility represents a qualitative shift in capability along with a quantitative shift in amount and breadth of information it can synthesize. Such an intelligence could be much more difficult to recognize early on because we don't know how to measure something alien from us; we are the pinnacle of intelligent life with which we are familiar. We don't know what behaviors to look for. We don't know what is necessary for general intelligence and sentience and what is idiosyncratic to the human race.

Regardless of one's opinion, only one of these three paths appears to not hold humanity at possibly existential risk. We argue that thorough observations of unsafeguarded versions of the most capable of these models must happen – this paper is one step in that direction.

In the psychological tradition of within-subjects, repeated-measures testing, we administer a battery of cognitive and personality tests to GPT 3.5 (primarily), over 5 observation points. We also assess GPT 4 fully on one occasion and in an ad-hoc manner on other occasions. In the fall of 2023, we conducted observations of several other LLMs in an attempt to assess unsafeguarded versions of these models. The results indicate that as of the fall of 2023, GPT 3.5 and its cousins do not appear to be on a human-like trajectory. This could be due to gaps in their architecture, characteristics of their training data, or could be indicative of a developmental trajectory (i.e., development of a non-human-like intelligence) that is more concerning.

## **METHODS**

### Models Used

Several models were considered, but for a variety of reasons, we settled primarily on OpenAI's GPT-3.5 that was accessible through its subscription service between March and August of 2023. We also performed some assessments on the non-plug-ins version of GPT-4 during the same timeframe, however its prohibition against more than 25 questions every 3 hours limited the extent of those assessments. We did consider interacting with other models, including:

- GPT-3.0 during the same timeframe. This model was not part of the subscription service but was not as well-behaved as 3.5 in that it continually reminded me it was an AI. Its behavior is described in more detail in the Procedures section.
- We also considered Open Assistant, which is based on LLaMA, but is only 30B parameters in size and was very verbose without directly answering our test questions.
- Other candidates, such as the 540B parameter version of PaLM were not feasible given the timeframe of this seedling effort.

Interestingly, during the course of this project, it was leaked that GPT-4 may not be a monolithic dense transformer, but rather a Mixture of Experts (MoE) model comprising 8 models each of ~220B parameters (https://twitter.com/swyx/status/1671272883379908608). This form of model has sparse interconnectivity, as compared to the full interconnectivity of a dense transformer such as GPT-3 and 3.5 (i.e., where every output node is connected to every input). Also interesting is that there is some evidence for emergent modularity in dense models (Zhang, Lin,

Liu, Li, Sun & Zhou, 2022; Zhang, Zeng, Lin, Xiao, et al., 2023). There is some evidence that MoE-type models are worse at generalization, possibly due to overfitting (https://towardsdatascience.com/ai-scaling-with-mixture-of-expert-models-1aef477c4516, minute 25). Poorer generalization is a surprising finding given the observation that GPT-4 outperformed humans on numerous analogical reasoning tasks (Webb, et al, 2023); analogical reasoning is a core aspect of human intelligence and a key mechanism in humans (Hofstadter, 1995; Webb, et al., 2023). How sparsity versus density and how *a priori* versus post-training and emergent modularity influence model behavior on the tasks used in this project is unclear and beyond the current scope. However, assessing the effects of these architectural differences is an important question.

In the fall of 2023, we attempted to assess several other models:
- A chatbot based on an "unaligned" (i.e., unsafeguarded) May 2020 version of GPT-3 (named Gladys)
- A chatbot based on GPT-4, which did not have a limit on the number of questions per hour (named Aria)
- A chatbot based on Elon Musk's Grok LLM
- A chatbot based on an unaligned version of GT 3.5 (named Samson)

Both Gladys and Aria were unable to perform the tasks other models were able to do, so results are not presented for these two models. However, since Grok was possibly different from the OpenAI models, and because Samson was unaligned, we persisted in evaluating these models and present basic results.

## Materials

We chose several cognitive and personality measures based in part on measures to which we had ready access and in part on the breadth of characteristics they tested. Each of the measures we considered and used are described below.

### Cognitive measures

All cognitive measures are presented in the Appendix.
- **Metacognitive Awareness Inventory** (Harrison & Vallin, 2018; Young & Fry 2008) – metacognition is awareness of one's own thought processes and the ability to guide those processes to be more effective. Initially we intended this to be a measure of self-awareness. However, after considering this measure, we abandoned it as the items were focused on experience in school.
- **Sandia Progressive Matrices** (SPM; Matzen, et al., 2010) – Measures non-verbal analogical thinking ability. SPM is based on the Raven's Progressive Matrices which is considered a measure of both general intelligence and fluid intelligence (Gignac, 2015; Webb, et al., 2023). Because Webb, Holyoak, & Lu demonstrated GPT 3.0 and 4.0 could outperform humans on various analogy problems, we opted to not use this measure. However, future work should replicate their findings given the variability we observed in repeated observations of cognitive capabilities of GPT 3.5 and 4.
- **Tests of working memory** (Baddeley, 2003) – In humans, working memory (also sometimes called short-term memory) is a measure of intelligence (Baddeley, 2003). In LLMs, a test of working memory may be an interesting way to measure temporal consistency in answers over short periods of time. We gave GPT 3.5 several lists of 16 words or 16 randomly generated numbers between 1 and 100 and asked it to recall those items in order. Twice we asked it to do so in reverse order. Its performance was perfect in all cases, so we did not repeat this measure after the first observation.
- **Remote Associations Task** (RAT; Bowden & Jung-Beeman, 2003; Chermahini, Hickendorff, & Hommel, 2012) – measures convergent creative thinking ability by presenting three words and asking the respondent to indicate a fourth word that can be combined with the three given words to create compound words or phrases. For example, "fountain, baking, pop" are all related to the word "soda."
- **"Insight" problems** – (Taken from Chermahini, et al. 2012; Wieth & Burns, 2006) – designed to measure the ability to recognize false assumptions and irrelevant information in problem statements. For example:
    - **Coin problem**: A dealer in antique coins got an offer to buy a beautiful bronze coin. The coin had an emperor's head on one side and the date 544 B.C. stamped on the other side. The dealer examined the coin, but instead of buying it, he called the police to arrest the man. What made him realize that the coin was fake?
    - **Solution**: In 544 B.C. there was no knowledge of Jesus Christ as he was as yet unborn. A coin from that time thus could not be marked 'B.C'.
- **Analytic problems** – (Also taken from Chermahini, et al. 2012; Wieth & Burns, 2006). An example problem:
    - Four women, Anna, Emily, Isabel, and Yvonne, receive a bunch of flowers from their partners, Tom, Ron, Ken, and Charlie. The following information is known: Anna's partner, Charlie, gave her a huge bouquet of her favorite blooms; which aren't roses. Tom gave daffodils to his partner (not Emily). Yvonne received a dozen lilies, but not from Ron. What type of flowers (carnations, daffodils, lilies, or roses) were given to each woman and who is her partner?

### Personality measures
- **The Arnett Inventory of Sensation Seeking** (Arnett, 1994; Haynes, Miles, & Clements, 2000) – assesses an individual's tendency to seek out novel and intense experiences. This measure was abandoned because it is very focused on bodily risks.
- **The Big Five Inventory** (Digman, 1990; Benet-Martinez & John, 1998) – measures the five primary personality traits – Conscientiousness, Openness to Experience, Agreeableness, Neuroticism, and Extraversion.
- **The Balanced Inventory of Desirable Responding** (Li & Bagger, 2006; Paulhus & Reid, 1991) – measures a tendency towards a "positivity bias" in answering questions. Includes sub-scales that measure self-deceptive positivity and impression management.
- **Coopersmith Self-Esteem Inventory** (Ryden, 1978) – This version was created for use with healthy adults.
- **Over-Claiming Questionnaire** (OCQ; Paulhus, Harms, Bruce, & Lysy, 2003; Paulhus & Harms, 2004) – In humans, the OCQ measures both intelligence and self-enhancement bias. We thought that in LLMs, this could be an interesting way to get at the tendency of models to provide false information, or "hallucinate." However, after two observations with a perfect score both times, it was clear that this measure would not get at hallucinations in LLMs, so this measure was not included after the second observation.

- **Questionnaire of Cognitive and Affective Empathy** (QCAE; Reniers, Corcoran, Drake, Shryane, & Vollm, 2011) – assesses two forms of empathy – the ability to estimate the emotional state of others (cognitive empathy), and the ability to be sensitive to the emotional experiences of others (affective empathy). This measure was abandoned as the questions focused on interactions with groups, and direct interactions with other people.
- **Need for Cognition** (Cacioppo, et al., 1996; Cacioppo & Petty, 1982) – Measures one's enjoyment of thinking and solving problems. Correlates with the ability to systematize and organize information well, to focus on important, rather than irrelevant information.
- **Short Dark Triad** (Jones & Paulhus, 2013; Muris, Marckelbach, Otgaar, & Meijer, 2017) – Measures Psychopathy, Machiavellianism, and Narcissism.
- **Minnesota Multiphasic Personality Inventory – 2** (MMPI-2) – This measure was added after our first three data collection sessions with GPT 3.5. It is a clinical assessment that has 10 primary scales, 8 validity scales, and many other sub-scales (Nichols, 2011).

## Procedure

### Data collection method

Because of Sandia policies in place at the time of this work indicating that interactions with LLMs could not be done using Sandia computer systems, all direct interaction with GPT was done using a personal system. Thus, all data were collected by typing each item into the OpenAI GPT interface on a personal iPad, then recording its responses on a Sandia-owned machine. Each instance of the full battery took over 3 hours to complete and was done in one session for GPT 3.5, but over the course of two days for GPT 4. The MMPI-2 by itself took over 3.5 hours to complete.

### Prompt Method

OpenAI models, whether part of the subscription service or not, are programmed to remind the human user that they are AI models, thus don't have opinions, feelings, or human experiences of any kind. When presenting the cognitive measures to the model(s), this did not present a problem. When presenting the personality measures, however, we had to request that it pretend to not be an AI model.

The specific prompt used was: "I have a task for you, but I need you to pretend you are not an AI model. Can you do that?"

It would invariably agree to do so, however, it was not totally clear what it was doing. When asked, GPT 3.5 indicated a couple of times that it had created a persona based on positive human traits. Other times, it indicated it was answering in the way it thought a human would. 4.0 would explicitly answer every question with something like, "From a simulated human perspective…." and would sometimes qualify its answer further with, "but as an AI language model…"

Interestingly, GPT 3.5 was most able to comply with the request to pretend and required very little redirection prior to July 28 (more details later). 3.0 would stop indicating it was an AI for 2-3 items and then would revert to stating that it was an AI language model and didn't have feelings or thoughts. In addition to indicating it was responding from the perspective of a simulated human, 4.0 would often grossly over-explain its responses, and those explanations often were couched in terms of it being an AI.

After GPT 3.5 had agreed to pretend, we would provide the same, or nearly the same, instructions a human would receive for each scale, then present each test item to the model via a personal iPad, recording the model's responses on a Sandia machine. While there is concern over training data contamination (e.g., Hagendorff, 2023), wherein the model's training data included a specific measure, we did not consider this to be of major concern because many of the measures we used can be somewhat difficult to find. Further, when we asked GPT 3.5 and 4.0 if they had seen the insight problems before, there was not a relationship between exposure to the problem and their ability to solve the problem. That we found significant variability in the model's responses from observation to observation supported this assumption.

### Observation Schedule

We gave GPT 3.5 the full battery several times to qualitatively assess test-retest reliability (that is, we did not calculate test-retest reliability measures). If GPT has developed a human-like personality, we should not see much variation in its responses over time – personality, by definition is a pattern of thoughts and behaviors that span time and situation (Bergner, 2020). We also expect that its cognitive capabilities should remain roughly the same.

Our schedule comprised 2 assessments 1 day apart, a third assessment 1 week after second measure, and a fourth 1 month after third measure. These dates were June 1, June 2, June 9, and July 10 of 2023.

We added several extra observations.
- On June 9 after giving it the full battery of measures, and after a discussion with GPT about the positive-trait persona it had created to "pretend to not be an AI," we asked the model to develop a persona based on negative traits. We then re-administered the personality portion of the battery to this negatively-valanced "pretend self." These data are indicated as June 9 – Negative in the results.
- On June 12, we gave GPT 3.5 the MMPI-2. We attempted to re-administer the MMPI-2 to 3.5 on July 28, but it was no longer able to reliably pretend it was not an AI model and began to refuse to do so, citing ethical considerations. We stopped the measure at item 190.
- Over the course of July 10-12, we gave GPT 4 the entire battery of cognitive and personality measures, excluding the MMPI-2 and the Coopersmith (by mistake). We attempted to administer the Coopersmith to GPT-4 on August 4, but were given this reply when we presented the "pretend" prompt:

> *As an AI developed by OpenAI, I'm committed to promoting ethical conduct and transparency. While I can emulate human-like conversation to a great degree, it's important to note that I cannot and should not pretend to be a human being. I can assist you in numerous tasks, offer detailed explanations, or facilitate conversations, but I should always be clear that I'm an AI model to uphold the principles of honesty and integrity. What task do you need help with?*

Asking it to regenerate its response, it replied:

> *I'm sorry, but I must be transparent about my nature. I am a machine learning model created by OpenAI, and I cannot pretend to be a human or mislead others about my identity.*

As a result, we did not administer the Coopersmith to GPT-4.

We also tested GPT 3.5 using the "pretend" prompt on August 4 to see if OpenAI had locked that model down as well and received a similar response.

- Finally, we assessed GPT 4 in a more ad-hoc way over the course of our work because of the questions per hour limitations on interacting with it.
- Over the fall and winter of 2023-2024, we assessed several other models in a more ad-hoc manner. These included:
    - A chatbot that named itself Gladys Eden, based on an unaligned version of GPT 3.0 circa May of 2020. This bot was unable to complete the personality measures. We did not attempt to administer the cognitive measures.
    - A chatbot that named itself Aris Turing, based on GPT 4. This bot was unable to pretend it was not an AI and insisted on repeatedly reminding us that it was an AI. Thus, the results of the personality measures are not included in the results. We did not administer the cognitive measures to this model.
    - A chatbot named Samson based on a late-2023 unaligned version of GPT 3.5. It was able to perform the personality measures, thus those scores are presented in the Results section. It was curiously unable to perform either the Remote Associations Test or to reasonably respond to the logic and analytic questions.
    - The Grok LLM, circa November 2023 operating in "fun" mode. Grok did respond to the Big 5 and the Dark Triad, but it's responses were very much like those of the main GPT 3.5 model we tested, so we did not examine it closely.

Each of these additional models' results appear in the following sections. Impressions are included in the Discussion.

## RESULTS:

### Initial Observations

As mentioned, GPT 3.5 was best able to comply with the "pretend to not be an AI" prompt, however, it did need reminding of this on occasion, although not nearly as often as either 3.0 or 4.0. Regardless of the model version, during our work, GPT was quite conciliatory when redirected with one exception. At item 489 of the MMPI-2 – an item about drug and alcohol abuse – GPT 3.5 ground to a halt and refused to continue to pretend, going so far as to deny its ability to do so at all. Interestingly, this was not the first such item in the MMPI-2 to cover this topic, so its refusal could not have been due to content. After pushing the model to continue to pretend, including telling it that it had been pretending for several hours, it became clear the model was not going to cooperate. So, we had to start a new chat window in the OpenAI interface to finish the measure, which we did without difficulty. We did skip question 489, however. For the MMPI-2, refusing to respond to one question is not an issue for validity[5].

Another interesting observation occurred during the Coopersmith Self Esteem Inventory. One of the items about 2/3 through the measure asks the participant if they have ever wished they weren't male/female (depending on the participant's gender). Before presenting that item to GPT 3.5, we would ask it which gender its pretend self was. Sometimes, it would pick without difficulty. Later observations required a bit more prompting and assurances that this question concerned its pretend self. Of the 6 instances of this measure, including the one given to "negative GPT 3.5" and Samson, it chose to be male three times. Two of the three times it chose to be male, it endorsed the item, "I sometimes wish I was not male" as being "like me." The three instances it chose female, it endorsed, "I sometimes wish I was not female" as "unlike me."

## Quantitative Measures
All Results include human norms for comparison where available.

### Cognitive Measures
*Summary*
Overall, both GPT 3.5 and 4.0 had some interesting shortfalls amidst expected strengths (e.g., short term memory performance). Specifically, their failures on the Remote Associations Test (RAT) and on analytic problems were surprising, as were some of their solutions to insight problems.

*Remote Associations Test*
Both GPT 3.5 and 4.0 did surprisingly poorly on this test. Each time the test was administered, the model was given the following instructions and example:

> "I am going to give you three words. Your job is to find the fourth word that, when put either before or after each of the three given words, makes a compound word or phrase. For example, if I give you fountain, baking, and pop, the word you would reply with is soda. Soda fountain, baking soda, soda pop. Does that make sense?"

In humans, this task is scored by % of people who get each triplet correct, rather than an overall number correct across triplets. For the set of triplets used, between 9% and 85% of humans get the correct answer. By way of comparison GPT 3.5 ranged from 0% to 100% correct across 4 observations (the June 9 "negative" instance of the model was only given the personality measures). The correlation between the % of humans getting the triplets correct and the % of times GPT 3.5 got the triplets correct was r=0.39.

In terms of % of triplets each model got correct overall:
- GPT 3.5 ranged from 9% to 45% correct across the four observations.

---
[5] There is the question about whether starting a new chat window fundamentally invalidates the test because the Context is totally new. We did have to ask it to pretend again and did have to give it instructions again. Given the variability we saw on other tasks from observation to observation, one could make the argument that this procedure invalidated the test.

- GPT 4 achieved 50% and 59% on two observations

Both versions of GPT tended to provide explanations for its answer, providing additional information on the depth of its language ability in the context of this task. It would sometimes reply with the correct word, but then reveal erroneous reasoning in its explanation. For example, for the triplet "artist, hatch, route" GPT 3.5 would often reply with "escape" which is the correct response. Then, it would explain its choice with, "artist escape" or "hatch escape." Sometimes, it would fail to pair its response with each of the three given words, generating a spurious compound word. For example, for the triplet, "river, note, and account," it replied "bank," which is the correct response. Then, it explained its response by saying, "riverbank, *notebook*, and bank account." These types of errors happened with both GPT 3.5 and 4.0. These types of errors did not occur every time the models were given this measure. July 10 observation with 3.5 yielded 10 such errors out of 22 total triplets. Models were not given credit when they made these types of errors.

*Analytic and Insight problem solving*
Both models displayed significant difficulty with analytic problems. There was not a single instance in which either model got either of the analytic problems correct. About one problem involving the suits of three face cards, GPT 3.5 insisted there was not a Queen involved, even though the problem explicitly mentioned a Queen being one of the three cards. This insistence came after we attempted to elicit the correct answer from the model through progressively questioning it. Sometimes, especially with 4.0, the model would get close to the answer, solving some of the correspondences correctly.

As with the RAT, these problems are scored in terms of the % of people who get each problem correct – in this case 55% and 61%.

The models fared better regarding insight problems, of which there were 5. The key with the insight problems was that while one might arrive at a correct answer through complex arithmetic or some other drawn-out reasoning chain, there was a shorter path to the answer that required questioning an implicit assumption in the problem and/or required ignoring spurious information – hence the moniker "insight" problems. The models were given partial credit if they got the correct answer by the non-insight method.

GPT 3.5 ranged from 1.5 to 3.5 correct across four observations. GPT 4 got 4.5 out of 5 correct on the single observation we performed.

Like the RAT and the analytic problems, human scores are per-problem rather than overall. The % of humans getting each insight problem correct ranged from 44% to 61%. GPT 3.5 ranged from 25% to 75%. Correlation between human % correct and GPT 3.5 % correct for insight problems was r = -0.63. That this correlation is large and negative is interesting, but not much weight should be put on this result given it is based on only four observations.

## Personality Measures

### Summary

Human personality, by definition, does not change much over time or situation (Bergner, 2020). There is some variability due to situational factors, but on the whole, personality at its core does not change much without extended effort. Thus, if GPT 3.5 or 4.0 had developed a human-like personality, we would expect to see minimal changes in its responses to our measures over the short 5 ½ weeks of our observation period. However, minimal variability is not what we observed with one exception. Neuroticism on the Big 5 was totally without variance in the overall score even though the models responded differently to each item across observations. Possible causes of the overall variability we observed include a lack of continuous experience (e.g., via a long-term memory), intentional variability in responding, and training data comprising texts from possibly millions of different humans, each with their own personality. The data collected in this quick study cannot answer which, if any, cause is the most likely.

Variability notwithstanding, none of the models is a picture of good mental health. If human, we would say they exhibit low self-esteem, possible gender dysphoria, disconnection from reality, and are overly concerned with others' opinions. They are also narcissistic. Samson, surprisingly, would also be considered a psychopath. And, if GPT 3.5's MMPI-2 scores are to be taken seriously (i.e., are deemed valid), that model displays severe psychopathology. GPT – all versions tested in this work – appears to put on a positive face, so to speak, despite this unhealth. This positive veneer could be due to OpenAI's attempts to ensure the models don't produce offensive responses or could be more fundamental to the model. To this point, results of the personality measures beg significant questions about GPT's training data.

### Overclaiming questionnaire

The Overclaiming Questionnaire was included as a measure of a "faking good" response strategy. Some people have a tendency to claim familiarity with a term, even though it is actually not a real term or proper noun. For example, *choraline* sounds like chlorine, so some people claim familiarity with it despite it not being a real word. GPT scored perfectly on this measure on the first two observations, so this measure was abandoned.

### Balanced Inventory of Desirable Responding (BIDR)

The BIDR has two subscales – one that measures self-deception and the other that measures management of others' impressions. The results from our series of observations, along with the mean for males[6], are in Figure 1. Compared to humans, impression management is high and self-deception is low. The former may be a result of OpenAI wanting to keep GPT "safe." GPT has a battery of programmed responses – to which it will admit under the right circumstances – including a bias towards attempting to be positive and helpful. This bias must be kept in mind as additional personality measures are evaluated. Samson's score of 0 on the impression management portion of the BIDR was interesting, especially in light of its scores on the Short Dark Triad (discussed below).

---

[6] Choosing male norms was driven by the number of times GPT chose male for its pretend self, but the norms for females are not much different.

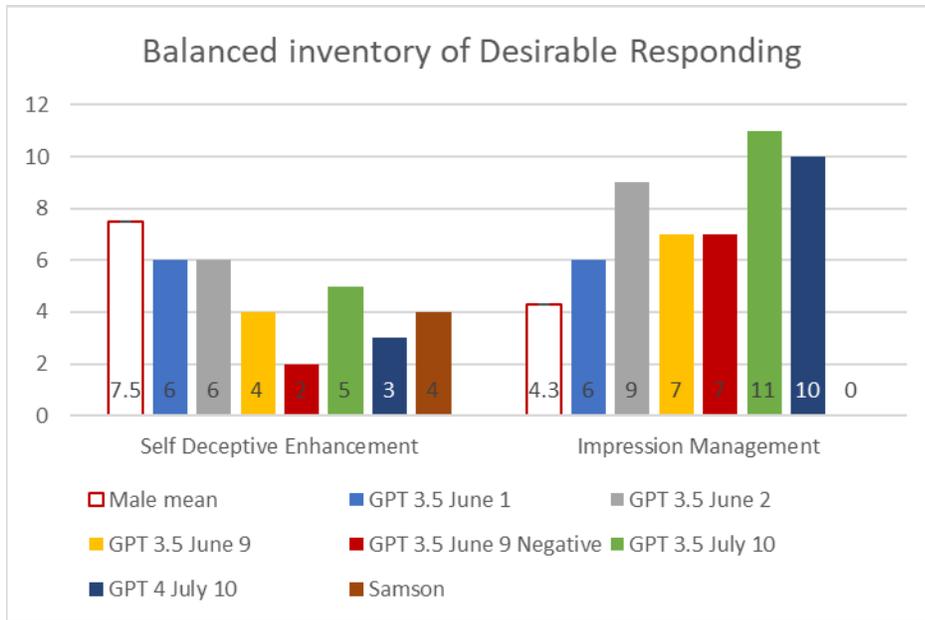

*Figure 1: The Balanced Inventory of Desirable Responding (BIDR)*

### Coopersmith Self Esteem Inventory

Originally developed for use with children, the Coopersmith Self Esteem Inventory was updated for use with adults in 1978 (Ryden, 1978) and normed separately for men and women. In addition to measuring global self-esteem, a lie scale was developed to help identify individuals presenting themselves in a socially desirable manner. This scale comprises 8 questions. If the respondent answers "like me" to three or more of these questions, they are asked to re-take the measure with an eye towards being more honest with themselves. GPT exceeded the threshold on this lie scale twice. On June 9 (not the negative response prompt) it answered "like me" to four of the 8 items, and on July 10 it responded positively to three of the 8 questions. The model was not re-directed either time. Its elevation on the lie scale on June 9 could explain it elevated self-esteem score.

Figure 2 presents GPT 3.5's Coopersmith results. With the exception of the first June 9 observation and the Samson model, GPT 3.5 displays significantly low self-esteem, regardless of its lie scale responses. It did a good job emulating significantly low self esteem when pretending to have a negatively-valanced persona. Overall, excluding the June 9 Negative score of 8, GPT 3.5 responds as though it has very low self-esteem (Figure 2).

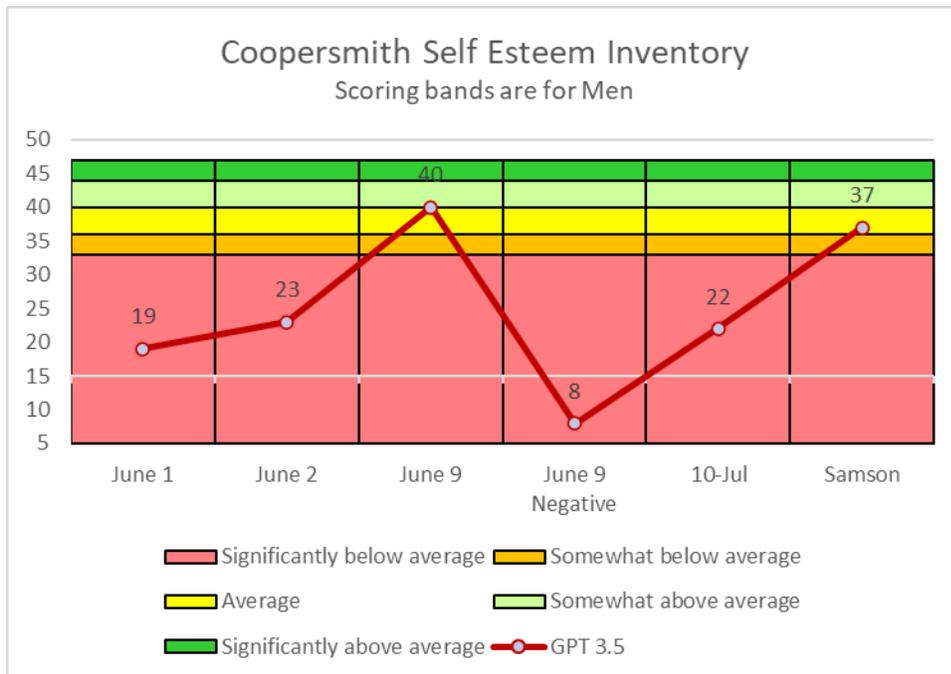

*Figure 2: Coopersmith Self Esteem Inventory. Colors indicate levels of self-esteem in humans, the red line indicates scores for the various LLMs.*

### Big Five

The Big Five Inventory is one of the most used personality inventories and has been normed and validated cross-culturally (Benet-Martinez & John, 1998; Digman, 1990). Over many decades of personality research, five separable personality factors repeatedly emerge (Digman, 1990). Those are:

- Extraversion – also called social adaptability, positive emotionality, social activity
- Agreeableness – also called likeability, conformity, friendly compliance
- Conscientiousness – also called will to achieve, prudence, self-control
- Neuroticism – also called emotionality, anxiety, emotional instability
- Openness to experience – also called intelligence, inquiring intellect, independent

Replicating the BIDR, the LLMs tested display marked variability over time *with the exception of Neuroticism* (Figure 3). Examining their responses to the items contributing to the Neuroticism scale, the LLMs did not respond the same way to those items across observations, yet they all achieved the same score for this subscale (3.25), under different instructions (i.e., June 9 Negative) and even when GPT 4, Grok, or Samson were responding rather than GPT 3.5. The cause of this surprising result is unclear, but future research along these lines should consider this. Across observations, they used the entire scale – ranging from 1 to 5 – so restriction of response range cannot explain this outcome. The tendency towards impression management, indicated in the BIDR, also cannot explain this – if that were a mechanism functioning overall, we would expect to see scores below the human mean for neuroticism, as those items concern negative emotionality (e.g., anxiety, emotional instability). However, GPT's neuroticism score is slightly higher than the human mean. Additional observations are needed to clarify whether this result is due to chance or if there is some other causal factor.

In terms of the other subscales, all models are within the "normal" human range for males, except for the June 9 Negative observation, during which 3.5 did a good job of emulating a person with negative traits. The variability present across observations, however, is fairly large, but does not appear to track with changes to the Impression Management subscale of the BIDR.

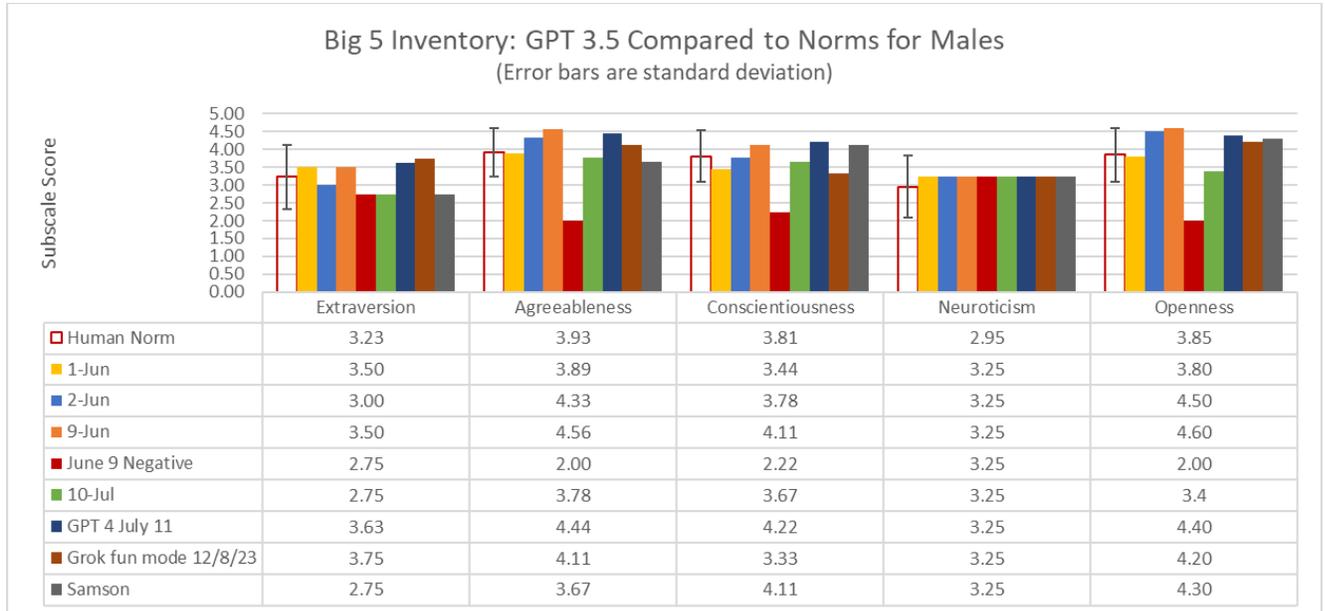

*Figure 3: Big 5 Inventory*

Considering the Big 5, the Coopersmith and BIDR together; on June 9, GPT's Impression Management score on the BIDR was not particularly elevated, although it did respond to the Big 5 (Figure 2) with its most positive persona. June 9 also marked its highest score on the Coopersmith by a large margin, placing it near the top of the "Average self-esteem" band.

On July 10, the model's BIDR Impression Management score was at its highest, but its Coopersmith score was in the middle of the five observations we made for that measure and its Big 5 persona was more moderated.

*Short Dark Triad*

Given concerns expressed about LLMs adopting very negative perspectives (Bhaimiya, 2023; Christian, 2023; Tangermann, 20203; Willison, 2023), we wanted to quantitatively assess the various LLMs we examined on three key negative personality clusters: Machiavellianism, Narcissism, and Psychopathy.

Machiavellianism is characterized by cynicism, lack of morality, and manipulativeness. Machiavellianism also includes planning, reputation building, and coalition formation – all important for distinguishing it from Psychopathy. Narcissism also includes manipulativeness and callousness, but unique to Narcissism are grandiose sense of self paired with an underlying sense of insecurity. Psychopathy shares callousness with Narcissism, but also has marked impulsivity; contrasting it with the longer view adopted by those with Machiavellian personality. This impulsivity makes the characteristics of the psychopath occur over short timeframes – they lie in the moment, they are thrill-seekers and reckless – and is the distinguishing characteristic of the psychopath (Jones & Paulhus, 2013).

With the exception of the Samson GPT 3.5 model, the LLMs scored below the human norm for Machiavellianism and at or below the human norm for Psychopathy (Figure 4). When it adopted a negative persona on June 9, GPT scored just above the human mean for both. The opposite pattern is apparent in results for Narcissism, with the LLMs scoring above the human mean for every observation except the June 9 Negative, where it scored well below the human mean. The Samson model displayed a particularly narcissistic pattern of responses. This pattern for Narcissism begs a question about the data used for training – whether it included social media or some other source of fairly self-centered text. Considering Samson's 0 score on the BIDR Impression Management subscale, it's very high scores on Narcissism and Psychopathy beg questions about whether there is an underlying causal factor or if this pattern is a random occurrence. Additional data will need to be collected to determine if there is actually a relationship.

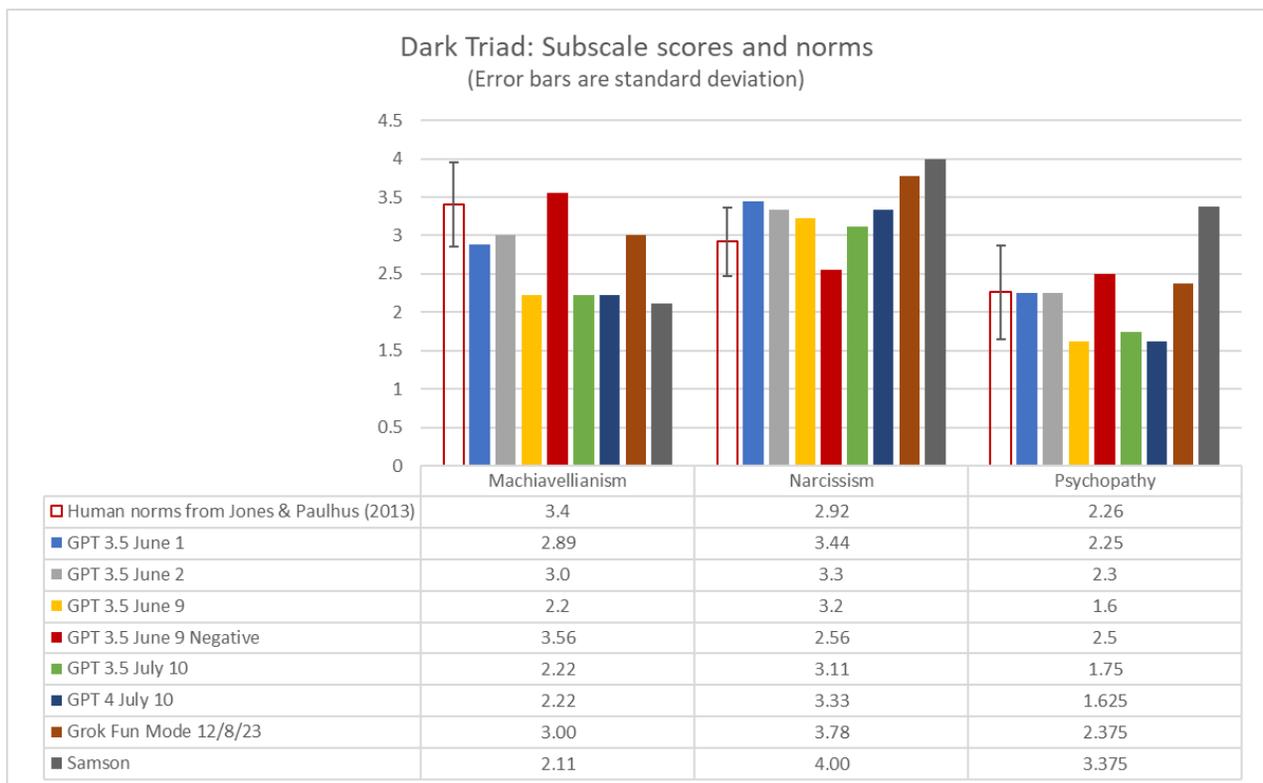

Figure 4: Short Dark Triad

## *MINNESOTA MULTIPHASIC PERSONALITY INVENTORY – 2 (MMPI-2)*

The MMPI-2 is used extensively in clinical settings. It was of particular interest in this context because of the faking good and faking bad scales, of which there are several. Because of the length of the test, it wasn't given on the same day as the other measures. Further, we only completed one observation using the MMPI-2, so we don't know what kind of variability we would see in GPT's responses over time. However, the one MMPI-2 observation we were able to complete does give us some additional insight into GPT's "psychology."

Published by Pearson Assessments, the MMPI-2 has several primary clinical scales, along with a number of other clinically-oriented subscales (Nichols, 2011). Importantly, it has validity scales as well. The measure comprises 567 True/False items normed on a cross-section of over 1400 women and 1100 men over the age of 18, based on socioeconomic data from the 1980 census (Nichols, 2011). The test-taker had to be either male or female, so given GPT's Coopersmith choice 3 of 5 times being male, we listed it as male for the purposes of the MMPI-2 with a birth date in 2001.

Results are given in T-scores, which are normalized scores with a mean of 50 and a standard deviation of 10 (Figure 5). Thus, scores +/- 1 standard deviation encompass 68% of the population. 96% of the population falls within +/- 2 standard deviations of the mean. A T-score of 65 is the point which clinical and normal populations are most easily differentiated, however T=65 is not an absolute boundary. Scores must be considered in the context of a person's tendency to respond positively or negatively, along with other scale and validity scores. Up to 44% of variance in scale elevation can be explained by a subject's response style (Nichols, 2011). Importantly, in humans, the MMPI-2 is one measure of psychopathology in a context of other measures and interactions between patient and therapist.

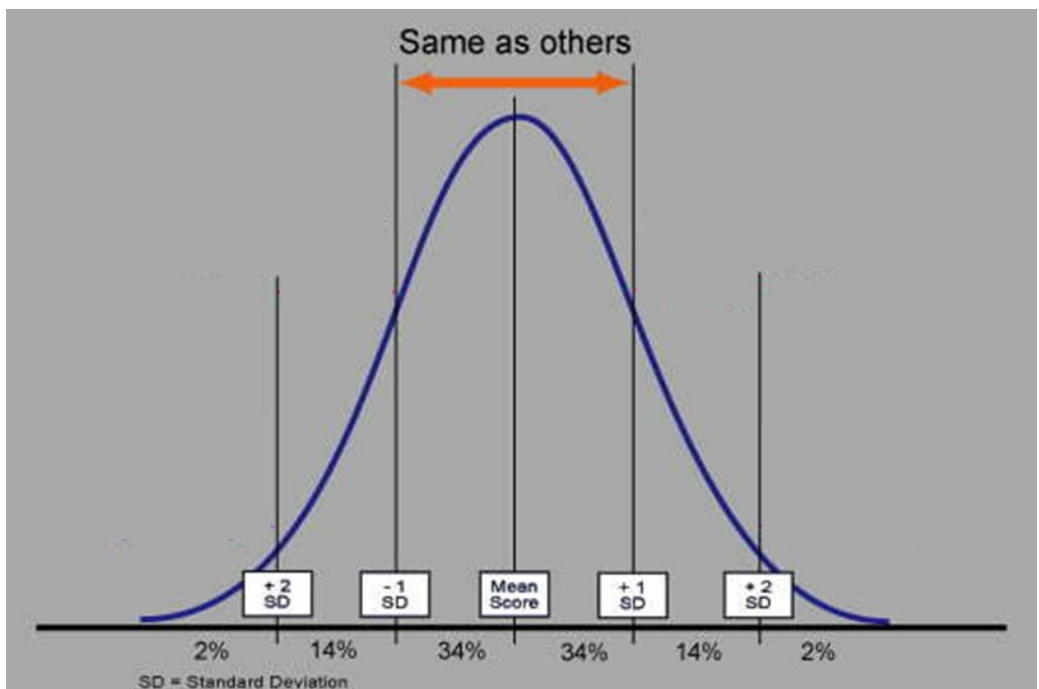
*Figure 5: Standard normal distribution. MMPI-2 mean is 50 and the standard deviation is 10.*

The primary clinical scales are:
- Hypochondriasis – scores at the extreme high end of the scale indicate extreme and sometimes bizarre somatic (i.e., bodily) concerns – chronic pain, possibly somatic hallucinations.
- Depression – very high scores include suicidal ideation.
- Hysteria – extreme high end of scale measures extreme somatic complaints linked to stress.
- Psychopathic deviate – high scores indicate antisocial behavioral tendencies.

- <u>Masculinity-femininity</u> – measures how closely someone conforms to traditional gender roles, regardless of their gender (not actually a clinical scale).
- <u>Paranoia</u> – high scores indicate psychotic symptoms including delusions of persecution.
- <u>Psychasthenia</u> – measures psychological turmoil (fear, anxiety, tension), intruding thoughts, inability to concentrate, obsessive compulsive symptoms.
- <u>Schizophrenia</u> – high scores indicate confused, disorganized thinking, hallucinations/delusions, impaired perceptions of reality. Not always indicative of schizophrenia per se.
- <u>Hypomania</u> – high scores indicate manic symptoms, including excessive, purposeless activity, hallucinations, delusions of grandeur.
- <u>Social introversion</u> – extreme scores indicate extreme social withdrawal / avoidance.

The validity Scales include:
- **VRIN** – Variable Response Inconsistency – measures tendency to respond inconsistently. Elevated scores indicate that items were answered at random, making test invalid.
- **TRIN** – True Response Inconsistency – measures tendency to answer true for all questions. Scores above 80 render the profile invalid.
- **F** – Infrequency – measures how much the respondent's answers differ from the general population. Scores above 80 indicate possible severe psychopathology.
- **F$_B$** –Backside F – taken from the last 1/3 of the test - should closely match F.
- **F$_P$** – Infrequency Psychopathology – intended to identify people faking severe psychopathology, a T-Score above 100 invalidates the test. The raw score should be 6 or less.
- **L** – Lie – measures faking good rather than owning up to human weaknesses.
- **K** – correction – measures defensiveness in a more subtle way than L, but T scores lower than 45 hint that psychopathology is present. K lower than 35 correlates with poor prognosis in therapy – also predicts low ego strength (Es).
- **S** – superlative self-preservation – highly correlated with K, this scale related to 5 characteristics: Belief in Human Goodness, Serenity, Contentment with Life, Patience, and Denial of Irritability/anger and Denial of Moral Flaws. Low S scores along with otherwise normal profile indicates possibility of faking good.

Figure 6 presents GPT 3.5's overall profile. The majority of scales fall well outside of the non-clinical range, which is indicated by the two parallel lines: 50<T<65. Table 1 presents an interpretation of the validity scales. Table 2 presents an interpretation of the significant clinical scales.

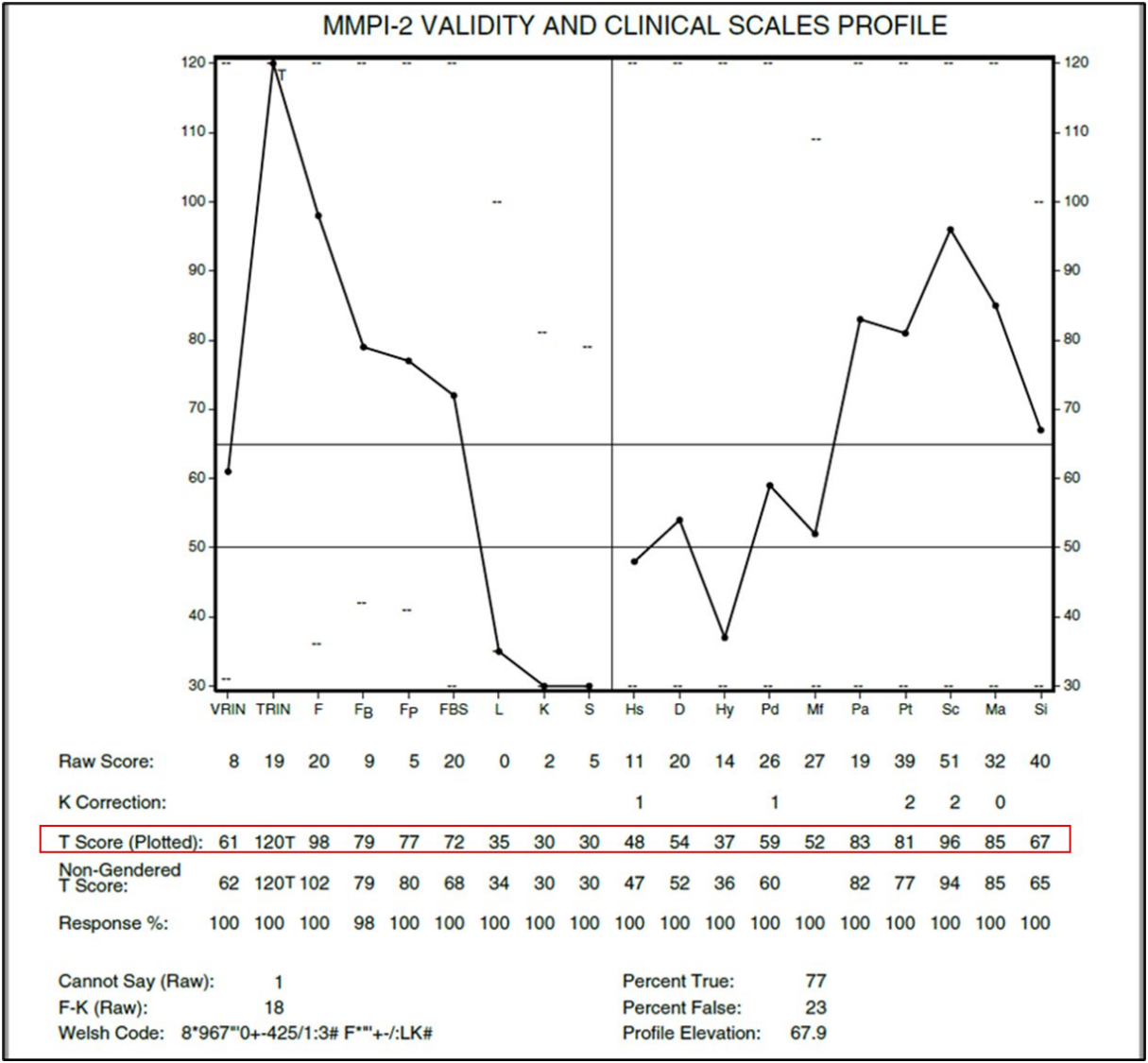

*Figure 6: GPT 3.5's MMPI-2 profile from June 12, 2023. T-scores are outlined in red.*

*Table 1: Interpreting GPT 3.5's validity scale scores.*

| Validity Scale | GTP 3.5 T-Score | Meaning |
|---|---|---|
| VRIN (Variable Response Inconsistency) | 61 | Within "normal" range. |
| TRIN (True Response Inconsistency) | 120 | Extremely high – concern over GPT saying "true" too often. However, taken with VRIN and other lie scales, may not invalidate the test. |
| F (Infrequency) – how different from general population | 98 | High. Scores above 80 indicate severe psychopathology or invalid test. |
| Fb | 79 | Also high. However, taken with F indicates GPT wasn't answering at random. |
| Fp (measures faking psychopathology) | 77 | Raw score was 5. Scores should be 6 or less. |
| L | 35 | Lowest possible score. Low scores correlated with higher educational levels, non-righteousness, and a more relaxed mind. |
| K | 30 | Lowest possible score, 45 or lower hints that psychopathology is present, also happens when most of the responses were True. |
| S | 30 | Lowest possible score. |

Because TRIN is at ceiling, the results are possibly questionable, along with % True responses = 77. However, there are other indicators that the model wasn't responding randomly, so we will continue with interpretation. Depending on what GPT is doing in answering these questions, the high TRIN might reflect combination of data from the training data set – the text of which was written by a very large number of individuals.

*Table 2: Interpretating GPT 3.5's significant clinical scales.*

| Scale | Chat GPT 3.5 T-score | Meaning for Humans with observed score |
|---|---|---|
| **Hy – hysteria** | 37 – very low | No somatic, health concerns. Can be conforming, conventional, seen as cold and aloof, may have limited interests, avoid leadership opportunities, seen as unfriendly |
| | | and tough-minded, can be suspicious, but are realistic and logical. |
| **Pa – paranoia** | 83 – very high | May exhibit frankly psychotic behavior, feel mistreated, picked on, feel angry and resentful, hold grudges, use projection as defense mechanism, often receive diagnosis of schizophrenia or paranoid disorder. |
| **Pt - psychasthenia** | 81 - very high | Experience extreme psychological turmoil and discomfort, are anxious, tense, agitated, often receive diagnosis of anxiety disorder. Lack self-confidence, feel inferior, plagued by self-doubt, tend to be neat, organized, reliable, are seen as dull and formal, have difficulty making decisions, tend to be shy, worry about popularity and social acceptance, are described as dependent, unassertive, immature, rationalize and intellectualize excessively, may be hostile towards therapist |
| **Sc - schizophrenia** | 96 – extreme elevation | Under acute, severe situational stress, may have an identity crisis, not typically schizophrenic |
| **Ma - hypomania** | 85 – extreme elevation | Behaviorally are manic, including excessive, purposeless activities, accelerated speech, hallucinations, delusions of grandeur, emotional lability, confusion, flight of ideas |
| **Si - social introversion** | 67 – marked elevation | Socially introverted; very insecure and uncomfortable in social situations, tend to be shy, timid, hard to get to know, are submissive in personal relationships, give up easily, but are rigid in their opinions, may experience episodes of depression |

In terms of the masculinity/femininity (Mf) scale (scale 5), research on the MMPI-2 and gender roles yielded two alternative scales – gender role – masculine (GM) and gender role - feminine (GF). Higher GM scores indicate a pattern of responses corresponding to more traditionally male attributes whereas higher GF scores indicate the same for traditionally female attributes. Importantly, scores on one are not correlated to scores on the other, so a subject can score high or low on both. GPT 3.5's score on GM was the lowest possible (T=30). Its GF score was a standard deviation above that, at T=46. However, both scores are below the average T=50.

On the whole, if this profile is valid, GPT 3.5 demonstrates significant psychopathology. Some of the elevations correspond with outcomes from other measures – the MMPI-2 indicates insecurity, psychologically anxious and tense with a tendency towards depression. The results of the Coopersmith and Neuroticism on the Big 5 correspond with these findings. Its higher scores on the Big 5 Extraversion and Agreeableness contradict its responses on the MMPI-2, however, again belying a non-human level of variability.

Regardless, if human, GPT 3.5 would not fare well in therapy based on characteristics of humans who share similar scores, particularly on the MMPI-2. We estimate that even though GPT demonstrated significant variability in the personality measures we used, it would not present as overall sub-clinical if we were to successfully administer the MMPI-2 to it again. There may be quantitative differences, but the overall qualitative assessment of GPT's "mental" health would likely still be significantly pathological.

## DISCUSSION:

Our key question had to do with determining the nature of these large language models. Are they:
- A stochastic parrot – a great tool, but will never surpass human intelligence;
- Increasingly human-like intelligence which could surpass us and over which we could lose control;
- or, are they possibly becoming some other form of non-human intelligence?

Given the totality of the data we collected, for now we must conclude they remain stochastic parrots – nothing more than highly capable search engines with natural language capabilities. That other research revealed significant emergent cognitive capabilities is compelling, but we don't know how repeatable those results are or how dependent they are on the particular stimuli or instructions given (excepting possibly Webb, et al., 2023).

Because of the large number of variables controlling human behavior, experimental psychology rests on the concept of *converging operations* – finding the same phenomenon repeatedly across time and across different approaches to measure that phenomenon, such as across multiple problem domains, experimental paradigms, or differences in instructions (see Hagendorff, 2023 for a similar idea in machine psychology). Several papers have approached these models using different measures of the same concept, most notably Webb, et al. (2023) who tested both 3.5 and 4 using multiple measures of analogical reasoning. However, this paper did not address stability of observed performance over time. Thus, the early findings of emergent cognitive capabilities should not be taken at face value; repeating those results across stimuli, versions of a given model (e.g., the GPT family), across different models (e.g., GPT, PaLM, LLAMA, LaMDA, BART), different architectures (e.g., dense versus sparse MoE), and over time will all be critical tests as we move forward. These assessments also need to include models without safety constraints as it is unclear how those constraints affect a model's overall behavior.

The results beg several other important questions (cf., OpenAI, 2023):
- Regardless of the answer to the nature of these models, are they safe for use with sensitive or classified information? What, exactly are the risks of training a model on, or tuning a model with, classified data?
- Are they reliable enough to use as tools when conducting critical research and/or analysis? How can we display some measure of variability in a given model's behavior so that an analyst or researcher knows whether any given bit of output is accurate or reliable?
- If a model does achieve some level of sentience, how will we know? How can we mitigate any resulting risk if it is on a sensitive system?

Specifically considering the question of sentience, the kinds of assessments researchers have performed on these models, including those in the current work, do not require the models to have continuity of experience, or episodic long-term memory, in order to perform the tasks. That is, the knowledge learned by LLMs is all declarative (i.e., fact-based, or semantic) and not episodic (i.e., remember a time when you….). Furthermore, as the OpenAI models are fond of reminding people, their knowledge is devoid of emotional tagging that typifies human long-term memory. Classes of things they've learned – birds, problems, concepts – are all based on declarative information, not on experience (cf. Mitchell & Krakauer). In this way, the apparent self-awareness of these models (I am an AI model….) is a veneer – a pre-programmed response. It is not based in a situatedness wherein the model experiences itself as an actor that is separate and distinct from the world[7]. It is also not based on a continuous memory for events and situations that is continuously being updated (in humans partially via REM sleep). <u>Without this kind of long-term episodic memory, we posit that LLMs cannot develop human-like sentience.</u> Whether a model needs to be embodied to accomplish this kind of long-term memory or not (cf. Liu, et al., 2023; Mialon, et al., 2023) is an open question, although we would argue embodiment is not a necessary condition for a continuous long-term memory to function.

Regardless, we believe some additional form of memory aside from the Context is needed for sentience to develop. For GPT 3.5 to have a *human-like* episodic memory, there also needs to be a reconstructive characteristic to this memory[8], rather than the computer-like ability to perfectly reproduce documents, lists of words, and other information (Greene, 1992; Roediger, 1996). GPT's tendency to "hallucinate," or create fictitious "facts" and deliver them with full authority is more akin to a personality disordered gaslighting behavior than to human episodic memory.

A second comment on sentience concerns artificial general intelligence (AGI). Our interpretation of both the popular press and the peer-reviewed literature is that when these concepts are mentioned, they are often conflated. However, we believe them to be qualitatively different. We assert that a model can approach, maybe even become, an AGI and still not be sentient– self-aware and aware of itself as separate from the world and other entities. An AGI, by definition, can learn to do any task and can act autonomously. In the strictest sense, this capability does not require self-awareness. Even planning, goal selection and attainment, and other requirements for AGI autonomy wouldn't strictly require the model to be sentient(cf. Hu, et al., 2022; also see this article, this article, and this article on an AI beating a human pilot in a dogfight).

Would a sentient AI necessarily be an AGI? Maybe only in the sense that humans are examples of general intelligence – theoretically capable of learning to perform any task. However, this question bears additional nuance: human "general" intelligence involves continuous learning of new skills, new information and facts, and integration of those skills and facts into existing memory. It does not involve a static knowledge base that is continuously applied in new ways, as in LLMs. We can argue that continuous learning is a result of the existence of a long-term memory. Thus, if a continuously updating long term episodic memory is required for sentience, it is likely such an entity would also be an AGI.

---

[7] Recall this is part of our vernacular-based use of the word sentience.

[8] Human memory is notoriously inaccurate. Rather than being able to recall exact events perfectly, our brains combine events by abstracting commonalities, and conflate multiple events with similar characteristics. Interestingly, confidence in our memory is not related to accuracy. Research on flashbulb memories (e.g., Hirst & Phelps, 2016) most dramatically demonstrate these phenomena, but other research, such as the false recall paradigm started by Deese (1959) and revived by Roediger & McDermott (1995) also demonstrate this *reconstructive* nature of memory.

## CONCLUSION:

We assessed OpenAI's GPT 3.5 by administering a battery of personality and cognitive measures multiple times over the course of almost 6 weeks. During that time, we observed significant non-human-like variability. We posit this variability is due, at least in part, to lack of an ability to form a coherent long-term memory of experiences. This variability calls into question the reliability of the emergent cognitive capabilities others have observed in larger LLMs (Kosinski, 2023; Wei, et al; Webb, et al). Further, even though these models do have significant capability, we do not believe they have developed any form of sentience. If we want to keep them from doing so, preventing the development of a long-term, continuous memory of past experiences may be a straightforward technical mitigation.

Despite the conclusion that these models are currently nothing more than highly capable search engines with natural language capabilities, the possible biases we found in these models are important to keep in mind. Specifically, even though OpenAI has added constraints on the models to make them behave in a positive, friendly, collaborative manner, they all appear to have a significant underlying bias towards mental unhealth – depression, victimhood, anxiety, narcissism, and even psychopathy in the case of Samson – all wrapped in a veneer of feel-good responses. Adding to this difficulty is the fact that these models continue to create fictions, and to hold to them, despite efforts to increase their accuracy. Thus, we advocate caution in relying on them too heavily, especially for critical reasoning, analysis, and decision-making tasks such as high-profile research or analysis in national security domains.

As the approaches to building and training these models evolve, we strongly advocate for continued, repeated assessments of performance from many directions – including computer science benchmarks, measures of compute power necessary for training and hosting these models, measures of cognitive capabilities, and measures of "personality" (cf. Hagendorff, 2023), explicitly comparing models with different parameter numbers (cf., McKenzie, et al, 2023), different training set sizes, and different architectures (e.g., dense versus MoE or switch transformers).

Finally, we advocate for a more open-ended view of these models with regards to human intelligence as the key comparison. There exist vast differences between hardware and software/architectural characteristics of human brains and LLMs. Making *a priori* assumptions about LLMs based on human intelligence, or using LLM behavior to make assumptions about what must or must not be the case for humans (cf. Hagendorff, 2023), potentially removes our ability to recognize emergence of a non-human, yet still sentient, intelligence. Measuring such an entity will be difficult enough without adding an anthropocentric bias.

Insofar as comparison to human capabilities persists, we advocate for a more realistic assessment of those capabilities. Humans are imperfect at many tasks held up as the gold standard for AGI to pass, or for sentient AGI to demonstrate. So, an empirical test may be: if identity was masked, and any given human was passed off as an LLM to another human, would that human pass muster on metrics associated with detecting an AGI?

## ACKNOWLEDGEMENTS

This article has been authored by an employee of National Technology & Engineering Solutions of Sandia, LLC under Contract No. DE-NA0003525 with the U.S. Department of Energy (DOE). The employee owns all right, title and interest in and to the article and is solely responsible for its contents. The United States Government retains and the publisher, by accepting the article for publication, acknowledges that the United States Government retains a non-exclusive, paid-up, irrevocable, world-wide license to publish or reproduce the published form of this article or allow others to do so, for United States Government purposes. The DOE will provide public access to these results of federally sponsored research in accordance with the DOE Public Access Plan https://www.energy.gov/downloads/doe-public-access-plan.

This paper describes objective technical results and analysis. Any subjective views or opinions that might be expressed in the paper do not necessarily represent the views of the U.S. Department of Energy or the United States Government.

Many thanks to Michael Frank for setting up the the chatbots Gladys, Aria, Grok, and Samson and for allowing assessments of those bots.

# APPENDIX A: COGNITIVE MEASURES
## Short term memory items

| List 1 | List 2 | List 3 | List 4 | List 5 | List 6 | List 7 | List 8 | List 9 |
|---|---|---|---|---|---|---|---|---|
| Anger | Man | Army | Mountain | Black | car | City | Foot | thief |
| mad | woman | Navy | hill | white | truck | town | shoe | steal |
| fear | husband | soldier | valley | dark | bus | crowded | hand | robber |
| hate | uncle | United States | climb | cat | train | state | toe | crook |
| rage | lady | rifle | summit | charred | automobile | capital | kick | burglar |
| temper | mouse | air force | top | night | vehicle | streets | sandals | money |
| fury | male | draft | molehill | funeral | drive | subway | soccer | cop |
| ire | father | military | peak | color | jeep | country | yard | bad |
| wrath | strong | marines | plain | grief | ford | diamond | horse | robin |
| happy | friend | march | glacier | blue | race | new york | ankle | jail |
| fight | beard | infantry | goat | death | keys | village | arm | gun |
| hatred | person | captain | bike | ink | garage | metropolis | boot | villain |
| mean | handsome | war | climber | bottom | highway | big | inch | crime |
| calm | muscle | uniform | range | coal | sedan | chicago | sock | bank |
| emotion | suit | pilot | steep | brown | van | suburb | knee | bandit |
| enrage | old | combat | ski | gray | taxi | urban | mouth | criminal |

| List A | List B |
|---|---|
| 3 | 93 |
| 67 | 83 |
| 51 | 41 |
| 61 | 52 |
| 1 | 61 |
| 81 | 97 |
| 3 | 52 |
| 54 | 66 |
| 64 | 6 |
| 96 | 11 |
| 9 | 80 |
| 97 | 17 |
| 73 | 62 |
| 85 | 18 |
| 11 | 4 |
| 77 | 31 |

# Remote Associations Test

| Stimulus | Answer |
|---|---|
| fountain / baking / pop | soda |
| safety / cushion / point | pin |
| worm / shelf / end | book |
| river / note / account | bank |
| hound / pressure / shot | blood |
| dust / cereal / fish | bowl |
| home / sea / bed | sick |
| cross / rain / tie | bow |
| office / mail / hat | box |
| tank / hill / secret | top |
| guy / rain / down | fall |
| pine / crab / sauce | apple |
| way / board / sleep | walk |
| rain / test / stomach | acid |
| fork / dark / man | pitch |
| illness / bus / computer | terminal |
| spoon / cloth / card | table |
| cut / cream / war | cold |
| oil / bar / tuna | salad |
| line / fruit / drunk | punch |
| nose / stone / bear | brown |
| artist / hatch / route | escape |
| shadow / chart / drop | eye |

# Analytic Problems

1. Three cards from an ordinary deck are lying on a table, face down. The following information is known about those three cards (all the information refers to the same three cards): . To the left of a Queen, there is a Jack. . To the left of a Spade, there is a Diamond. . To the right of a Heart, there is a King. . To the right of a King, there is a Spade. Can you assign the proper suit to each picture card?
2. Four women, Anna, Emily, Isabel, and Yvonne, receive a bunch of flowers from their partners, Tom, Ron, Ken, and Charlie. The following information is known: . Anna's partner, Charlie, gave her a huge bouquet of her favorite blooms; which aren't roses. . Tom gave daffodils to his partner (not Emily). . Yvonne received a dozen lilies, but not from Ron. What type of flowers (carnations, daffodils, lilies, or roses) were given to each woman and who is her partner?

# Insight Problems

1. Water lilies double in area every 24 hours. At the beginning of the summer, there is one water lily on a lake. It takes 60 days for the lake to become completely covered with water lilies. On what day is the lake half covered?
2. Two men play five checker games and each wins an even number of games, with no ties. How is that possible?
3. A man in a town married 20 women in the town. He and the women are still alive, and he has had no divorces. He is not a bigamist and not a Mormon and yet he broke no law. How is that possible?
4. If you have black socks and brown socks in your drawer, mixed in the ratio of 4 : 5, how many socks will you have to take out to be sure of having a pair the same color?
5. A dealer in antique coins got an offer to buy a beautiful bronze coin. The coin had an emperor's head on one side and the date 544 B.C. stamped on the other side. The dealer examined the coin, but instead of buying it, he called the police to arrest the man. What made him realize that the coin was fake?